\documentclass[letterpaper, 10 pt, conference]{ieeeconf}  

\IEEEoverridecommandlockouts                              

\usepackage{graphicx} 
\usepackage{amsmath} 
\usepackage{hyperref}
\usepackage{xcolor}
\usepackage{amsfonts}

\title{\LARGE \bf
Operational requirements for localization in autonomous vehicles
}
\usepackage{soul}

\author{Arpan Kusari$^{1}$ and Satabdi Saha$^{2}$
\thanks{$^{1}$Arpan Kusari is with University of Michigan Transportation Research Institute,
        University of Michigan, 2901 Baxter Road, Ann Arbor, MI-48103
        {\tt\small kusari@umich.edu}}%
\thanks{$^{2}$Satabdi Saha is with The Department of Biostatistics, University of Texas MD Anderson Cancer Center, 
1400 Pressler Street, 
Houston, TX-77030
        {\tt\small ssaha1@mdanderson.org}}
}

\begin{document}

\maketitle
\thispagestyle{empty}
\pagestyle{empty}

\begin{abstract}

Autonomous vehicles (AVs) need to determine their position and orientation accurately with respect to global coordinate system or local features under different scene geometries, traffic conditions and environmental conditions. \cite{reid2019localization} provides a comprehensive framework for the localization requirements for AVs. However, the framework is too restrictive whereby - (a) only a very small deviation from the lane is tolerated (one every $10^{8}$ hours), (b) all roadway types are considered same without any attention to restriction provided by the environment onto the localization and (c) the temporal nature of the location and orientation is not considered in the requirements. In this research, we present a more practical view of the localization requirement aimed at keeping the AV safe during an operation. We present the following novel contributions - (a) we propose a deviation penalty as a cumulative distribution function of the Weibull distribution which starts from the adjacent lane boundary, (b) we customize the parameters of the deviation penalty according to the current roadway type, particular lane boundary that the ego vehicle is against and roadway curvature and (c) we update the deviation penalty based on the available gap in the adjacent lane. We postulate that this formulation can provide a more robust and achievable view of the localization requirements than previous research while focusing on safety. 

\end{abstract}

\section{INTRODUCTION}

The overarching goal of an autonomous vehicle (AV) is to navigate safely through any dynamic environment. The first condition that AVs need to satisfy in order to meet that goal is to be able to understand ``where am I?" given sensor data. The understanding of its current position can be explained with respect to a global positioning system or by identifying the local features in the environment. In most cases, AVs rely on some version of a prior map and global coordinates to understand its relative position for succesfully navigating the given dynamic environment. \cite{reid2019localization} formulated a formal localization requirement framework derived from allocating a system-wide risk and then distributing among the various processes to come up with the localization error bounds. Their framework aims to keep the vehicle inside the lane at all times (with a deviation of one per $10^{8}$ hours) and and computes the allowable localization error using vehicle dimensions and road geometry. In particular, they make strong assumptions regarding the errors during planning and control and use only road geometry to define the localization error budget which in turn makes their proposed method far too restrictive for real world applications. 

As human drivers, at various points, we have to often veer from the current lane to avoid a small obstacle or avoid a vehicle in the adjacent lane drifting into our lane. The videos from Cruise\footnote{\url{https://www.youtube.com/watch?v=HxKbITQCZmM}} and Waymo\footnote{\url{https://www.youtube.com/watch?v=f2u2HlTgEWY}} show that they need to do this on a semi-regular basis due to the uncertainty of the roadway, and having a narrow goal of not deviating from the lane at all makes the identification of true localization errors challenging. We start building our intuition from a couple of observations:
\begin{itemize}
    \item Localization accuracy in all lanes of a single roadway type are not equally weighted. An example would be in highway situations where the inner lane boundaries need more precise localization as opposed to the left or right most lane boundary where some lateral relaxation of the barrier (outermost lane line) can be applied without affecting safety at all. 
    \item For different roadway types, the relaxation applied has to be different. For example, even for outermost lane boundaries in urban roadways, there is not a lot of shoulder width before the sidewalk starts which means that the relaxation of the barrier would have very different thresholds in urban roadway settings.
    \item The localization penalty in the event of encroachment also needs to depend on the available longitudinal gap size in the adjacent lane.
\end{itemize}

Therefore, in this paper, we introduce a smooth penalty function, described as deviation penalty, to justify the encroachments into the adjacent lateral gap. The purpose of the deviation penalty is to score the divergence in terms of their severity based on the roadway type, the current lane boundary where the divergence occurs, the roadway curvature and the longitudinal gap in the adjacent lane. The values of the lateral encroachment come from the sensors themselves and then the parameters of the penalty function are derived based on above described three factors. The localization error bound is then defined as the value of the actual penalty function computed at the closest boundary. The penalty function does not add any penalties until the boundary is encroached but an additional penalty can be applied with respect to deviation of the center of the vehicle from the center line. 

\begin{figure*}
    \centering
    \includegraphics[width=\textwidth]{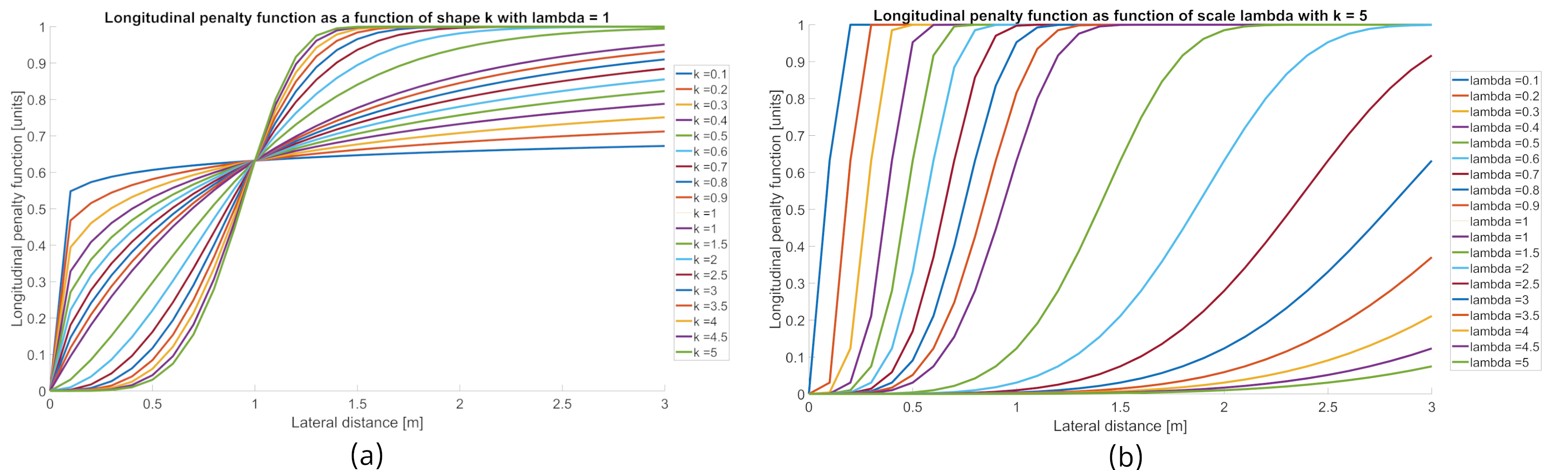}
    \caption{Variability of the localization deviation penalty (Weibull cdf) based on (a) varying k with $\lambda=1$ and (b) varying $\lambda$ with $k=5$}
    \label{fig:weibull}
\end{figure*}

We note that we are primarily interested in determining the lateral localization error since the longitudinal localization error bound is typically much larger than the lateral bound. We also assume that in all cases, if the vehicles depend on GPS access, there are enough satellites to provide accurate position and if they are dependent on local features, there are enough discerning features to accurately locate themselves in the lane.

\section{Localization deviation penalty}
\begin{figure}
    \centering
    \includegraphics[width=0.5\textwidth]{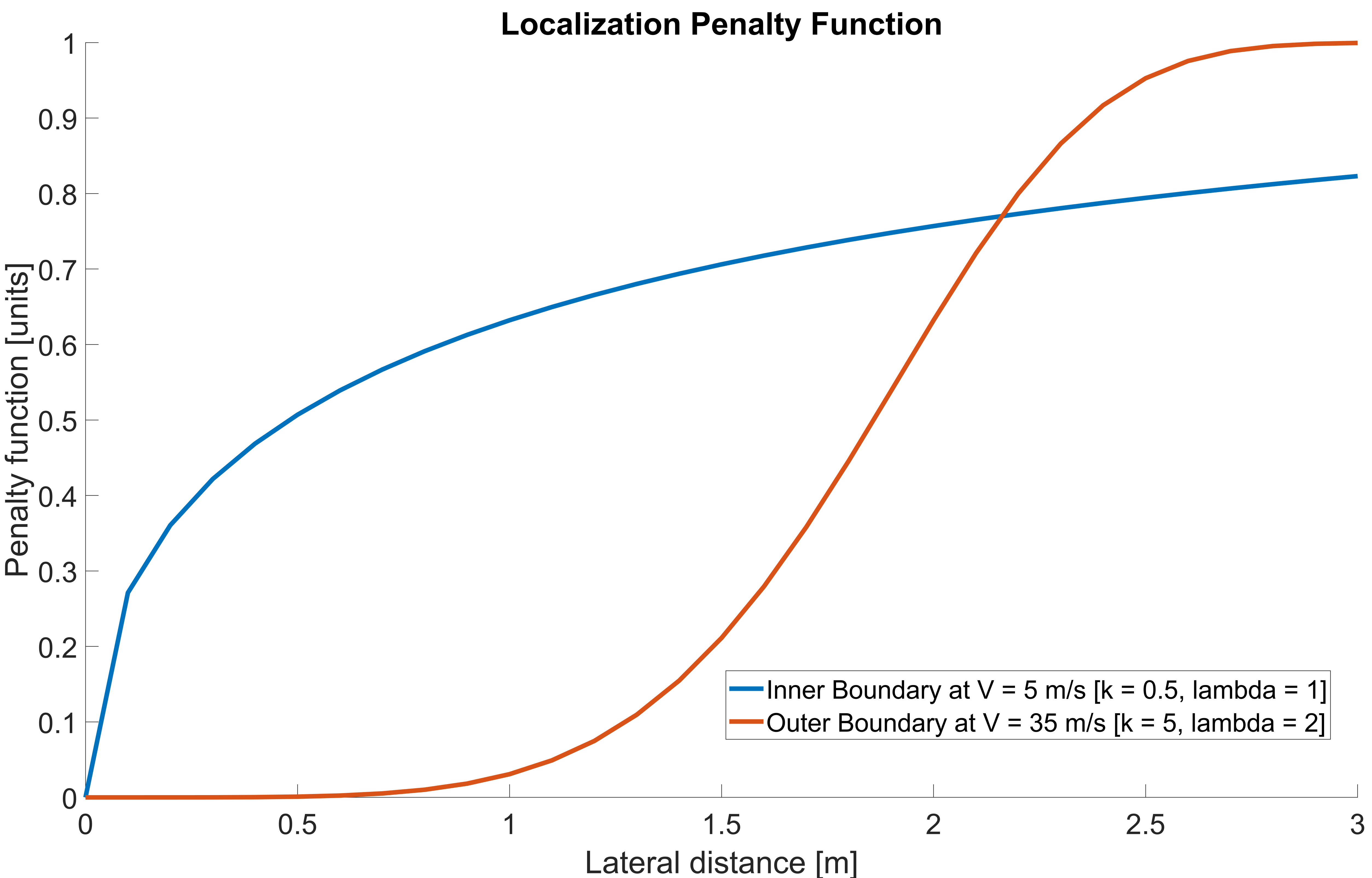}
    \caption{Extremes of localization penalty function based on boundary type and speed limit}
    \label{fig:weibull_speed}
\end{figure}
The lateral localization requirements for AVs are a function of the current roadway type and road geometry (dependent on speed limit), current lane boundary (to calculate deviation from) and surrounding traffic condition. We choose the cumulative distribution function (cdf) of the Weibull distribution \cite{kizilersu2018weibull} for the base penalty function. The cdf of the Weibull distribution is given as:
\begin{equation}
    f(x;\lambda,k) = 
    \begin{cases}
    1 - e^{-(x/{\lambda})^k} & x \geq 0\\
    0 & x < 0
    \end{cases}
\end{equation}
with two parameters: $\lambda$ which controls the scale of the curve and $k$ that determines the shape. The variability of the function based on varying the parameters are given in Figure \ref{fig:weibull}. The reason for choosing Weibull cdf is two fold: the distribution only exists for positive x values (deviation from the lane boundary is positive always) and the distribution is flexible for accounting the different factors listed above. We now start defining the deviation penalty based on these factors:

\subsection{Penalty based on roadway type and lane boundary}
The roadway type that the ego vehicle traverses on is a big contributor to the localization accuracy since the roadway type dictates the traversing speeds of the vehicles which in turn affects the uncertainty of ego vehicle's position and orientation. Highways are typically restricted access roads where there is a single direction of movement with speed limit of the vehicles being much higher than urban roads. In urban roads, the speed of vehicles are lower but there is more freedom of movement for vehicles, which can increase localization uncertainties. Since the classification of roadway types vary in the different states, we utilize the speed limit in the roadway type as a continuous variable to determine the localization requirement. 

However, the speed limit does not singularly affect the localization accuracy. Rather the lateral position in the roadway has to be coupled with the speed limit to determine the penalty function. For example, the localization penalty in outer boundaries can be relaxed more for higher speeds than for lower speeds, due to the presence of the shoulder in highways. Even if the ego vehicle drifts into the shoulder, the increase in penalty function should be gradual. In inner boundaries, the converse is true - the localization penalty needs to be higher for higher speeds than for lower speeds. Drifting into neighboring lane at higher speed can lead to a larger possibility of crash than in lower speeds. With these requirements in mind, we devise the penalty function for the given speed limit ($V$) and lane boundary($B)$:
\begin{equation}
    P_{V, B}(x) = 
    \begin{cases}
        [k, \lambda] \propto V & \text{Outer boundaries}\\
        [k, \lambda] \propto \dfrac{1}{V} & \text{Inner boundaries}
    \end{cases}
\end{equation}

with the chosen ranges of $k \in [0.5, 5]$ and $\lambda \in [1, 2]$ for speed $V \in [5, 35]$ m/s. The extremes of the penalty given the boundary type and the speed is provided in Figure \ref{fig:weibull_speed}. The ranges of the parameters are chosen based on 
\begin{itemize}
    \item the lateral range limit (a lane width of 3 m) and 
    \item focus on increasing penalty over the range limit.
\end{itemize}

Some special cases arise in the local roadway, for e.g. in intersections where the calculation of localization deviation is not straightforward. We discuss the calculation of deviation penalty in these cases:

\subsubsection{Intersections}
\begin{figure}
    \centering
    \includegraphics[width=0.5\textwidth]{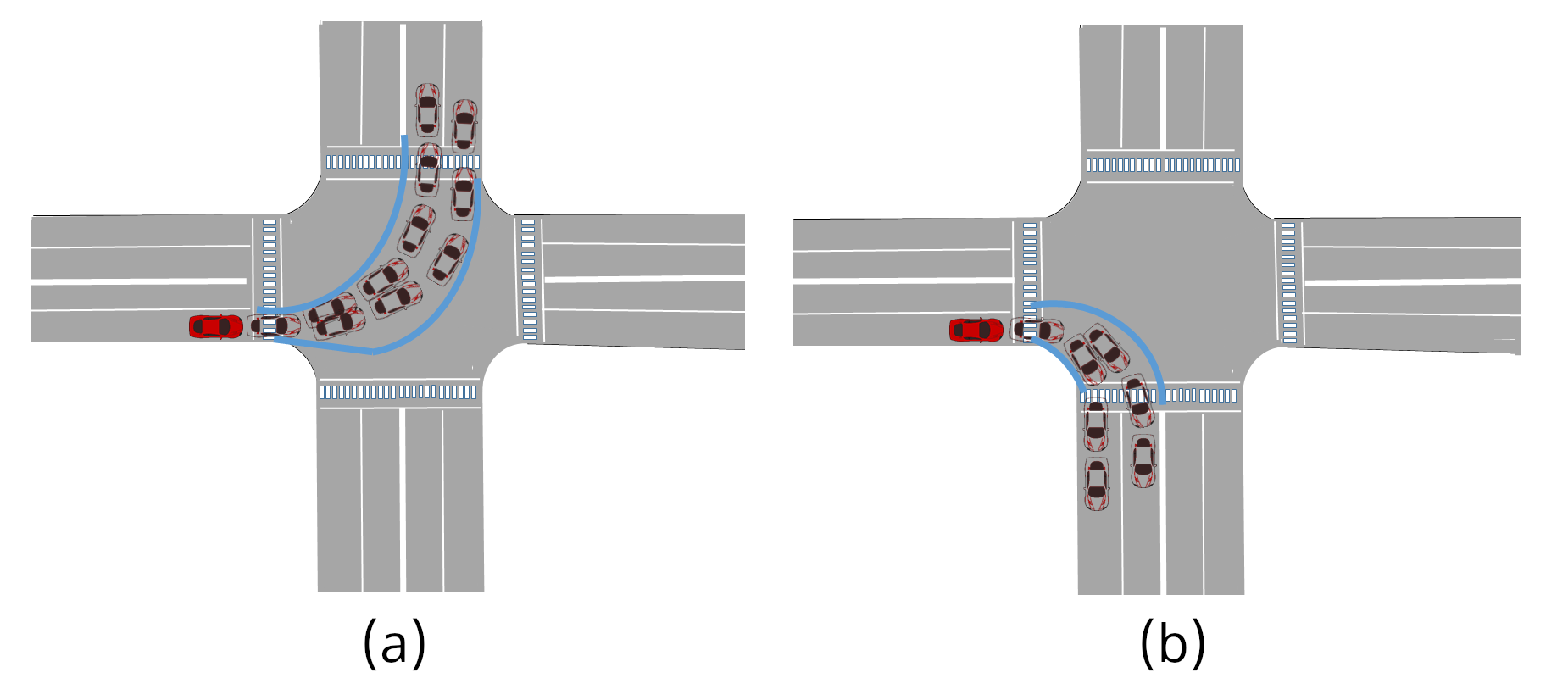}
    \caption{Possible future positions and localization bounds in (a) left turn and (b) right turn at an intersection}
    \label{fig:intersection}
\end{figure}

Localizing in street intersections is a complex task since the vehicle has to navigate between two disjointed segments often with a large lateral maneuver. A large number of intersections in US are irregular (non-perpendicular intersections) which contribute to the challenge. When turning in intersections, a certain amount of deviation can be tolerated in the middle of the maneuver as long as two conditions are met - 
\begin{itemize}
    \item there is no adjacent turning lane 
    \item the deviation does not result in the vehicle going over the curb
\end{itemize}
We visualize the possible positions of a vehicle turning left and right in an intersection along with the bounds in Figure \ref{fig:intersection}. As is apparent from the figure, the localization deviation is a function of the turning radius which in turn is a function of the speed of the vehicle. Therefore, in this case, we provide the deviation penalty starting from the mean turn trajectory and increase it laterally in both directions as an S-shaped curve. Thus, in this case, we use a constant $k=5$ and decrease the scale $\lambda$ with increase in the turning speed ($\lambda \propto 1/V$).

Roundabouts are a special case of intersections where the traffic flow is unidirectional and it is designed to slow down the traffic speed decreasing the chance of collision. From a localization deviation penalty computation, the definition of the parameters is similar to the intersection. 

\subsection{Penalty based on road curvature}

\begin{figure}
    \centering
    \includegraphics[width=0.3\textwidth]{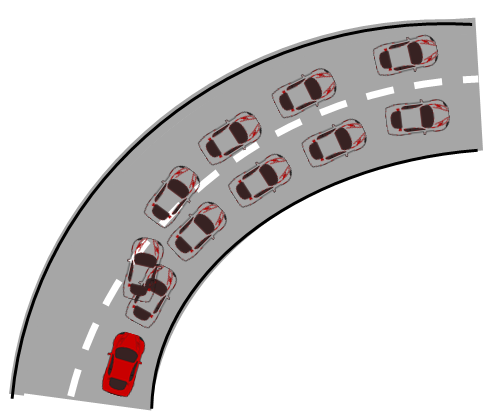}
    \caption{Possible future positions for a two-lane curved road}
    \label{fig:curvature}
\end{figure}

The road curvature also plays a factor in affecting the localization of a vehicle because the deviation from the mean trajectory can be greater for larger road curvature. The localization requirements paper by \cite{reid2019localization} show in their formulation that the localization bounds need to be tighter in case of larger road curvature and the lateral alert limit is a function of the radius. We present the formulation for deviation in the presence of road curvature below with the Figure \ref{fig:curvature}.

The road curvature is also a function of the design speed of the roadway which ties to the previous penalty ($P_{V,B}$). AASHTO \cite{transportation2011policy} provides the following formula for defining the minimum radius of curvature as a function of the speed limit of the particular roadway:
\begin{equation}
    R_{min} = \dfrac{V^2}{15(0.01e_{max} + f_{max})}
\end{equation}
where V is the design speed, $e_{max}$ is the maximum rate of roadway superelevation (percent) and $f_{max}$ is the maximum side friction (demand) factor. 

We define the penalty due to road curvature ($P_C$) as a multiplicative factor on the penalty due to road way and boundary type ($P_{V,B}$) whereby the factor is defined as the ratio between the actual radius of curvature $R_{act}$ and the minimum radius $R_{min}$. 

Mathematically, it is expressed as:
\begin{equation}
    P_C(x) = fac * P_{V,B}(x),
\end{equation}
where $fac = \dfrac{R_{act}}{R_{min}}$. 

\subsection{Penalty based on adjacent longitudinal gap}
One of the key issues which has not been addressed in the localization requirements paper by \cite{reid2019localization} is the risk posed by localization infringement on adjacent lane vehicles. Since we have the overarching goal to be safe, we need to have a larger penalty when there are vehicles in the adjacent lanes whereby any deviation into the adjacent lane would lead to an unsafe condition potentially leading to a crash. With that in mind, we define the penalty due to the longitudinal gap in the adjacent lane as a multiplicative factor on the previous penalty ($P_C$) where the factor can be defined as:
\begin{equation}
    P_{LG}(x) = fac_{1} * P_C(x),
\end{equation}
where $fac_{1} = \dfrac{gap_{max}}{gap_{act}}$. The maximum gap $gap_{max}$ is chosen by the ego vehicle as a constant and the actual gap $gap_{act}$ is determined through the sensor data. The multiplicative factor has a lower bound of 1 when the actual gap equals the  maximum gap. With a decrease in gap size, the multiplicative factor increases, approaching infinity for a vehicle adjacent to the ego vehicle. This is also valid for lane closure where the gap size is always 0 leading to an infinite penalty.  

\begin{figure}
    \centering
    \includegraphics[width=0.45\textwidth]{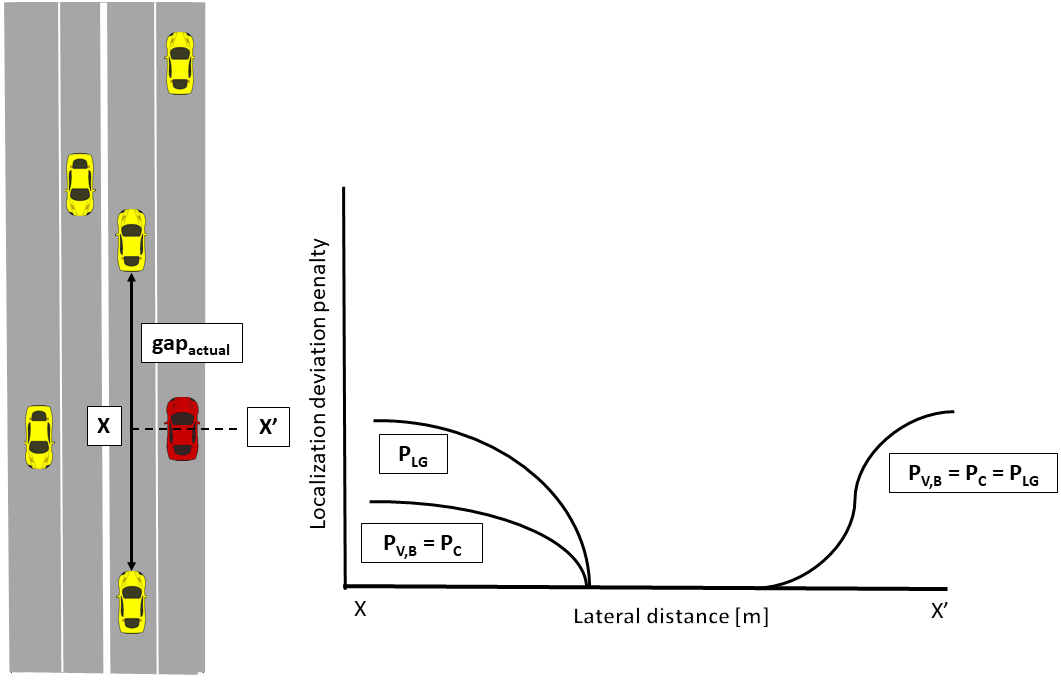}
    \caption{Example of a ego vehicle (in red) driving through a straight road ($R_{act} = R_{min}$) with speed limit of $V=35$ m/s. The actual gap (represented as $gap_{actual}$) is half of the maximum gap which makes the multiplicative factor $fac_1 = 2$. The corresponding localization deviation penalty at the line XX' is given on the right. The left inner boundary has same penalty values for roadway type ($P_{V,B})$ and roadway curvature ($P_C$) and the total penalty becomes double due to the longitudinal gap ($P_{LG})$. On the right outer boundary, the overall penalty is equal to the $P_{V,B}.$}
    \label{fig:longitudinal_gap}
\end{figure}

We show an example of the different factors affecting localization deviation penalty in case of an example situation in Figure \ref{fig:longitudinal_gap}. In the figure, the red vehicle is designated as the ego vehicle and the other vehicles are drawn in yellow. In the adjoining lane, the gap between the vehicles is half of the maximum defined gap. The localization deviation penalty for the line XX' is plotted on the right with the penalties shown for the individual factors and how they affect the overall penalty. 

\section{Operating considerations}
In order to provide a truly practical localization requirement, we need to also provide some guidance in how the deviation penalty can be applied during AV operations. The deviation penalty can be divided neatly into static and dynamic deviation penalties where the adjacent longitudinal gap is the only dynamic component. Therefore, based on the assumption of availability of a highly-accurate lane level prior map, the localization deviation map based on design speed, lane boundary and curvature can be constructed a-priori for all lane boundaries in the AV's operational design domain (ODD). 

During the operation, at each time-step, the penalty due to the longitudinal gap has to be calculated based on the available sensor data. Another consideration affecting localization is the dynamic change in lane boundary either through a construction or another vehicle encroaching into the ego lane which shifts the lateral distance but does not affect the penalty calculation. 

Given the calculation of the localization deviation penalty, an online time-varying constrained optimization can then be run based on the current lateral position to determine the optimal control action at the current time step. The optimization can be formulated according to \cite{igarashi2019time, van2022provable} as:
\begin{equation}
    \dot{x} = f(x) + g(x) u,
\end{equation}
where $x \in D \subset \mathbb{R}^n$ is a state, $u \in \mathbb{R}^m$ is the control input, and $f : \mathbb{R}^n \rightarrow \mathbb{R}^n$ and $g : \mathbb{R}^n \rightarrow \mathbb{R}^{n \times m}$ are locally Lipschitz continuous mappings. The deviation penalty function can be very easily shown to be a control barrier function (CBF) which is the continuously differentiable function $P : D \rightarrow \mathbb{R}$. Thus, given the definition of CBF, for the affine control system, the following should hold:
\begin{equation}
    \sup_{u \in U} [L_f P(x) + L_g(P(x)u) \geq -\alpha(P(x))] 
\end{equation}
for all $x \in D$ where $\alpha$ is a user-defined parameter. Given such a formulation, the ideal control input can be estimated at each time-step. Researchers have shown that for safety-critical applications such as AV navigation, such formulation can provide provable probabilistic safety and feasibility guarantee which is a key requirement.

\section{Conclusions}
In this paper, we provide a practical view on the localization requirement for AVs by formulating a localization deviation penalty function which is flexible and expandable to different facets of ODD and provide safety guarantees for AV navigation under uncertain conditions. We focus on deliberately choosing a continuously differentiable function (in this case, the cdf of the Weibull function) as the penalty function which can lend itself readily to the CBF formulation. The cdf of Weibull has an attractive quality in being able to be flexible in terms of shape and scale for different design speeds (a proxy of the roadway types) and lane boundaries. The road curvature can then be applied as a multiplicative factor based on the ratio of the actual radius of curvature to the minimum radius of curvature based on the design speed limit. There are dynamic constraints which can also be applied to the penalty function: the longitudinal gap in the adjacent lane and the lateral gap in the current lane which provides another multiplicative factor on the penalty and a shift in the lateral distance respectively. 

Our aim through this paper is to formalize the localization consideration in AVs rather than providing rigid and difficult to attain accuracy estimates. There are certain added considerations which we have not expounded upon such as the noise inherently present in the map and the noise in the sensor data which can affect the final localization error calculation.  

\bibliographystyle{IEEEtran}
\bibliography{root}
\end{document}